\documentclass[letterpaper, 10 pt, conference]{ieeeconf}  

\IEEEoverridecommandlockouts                              

\newcommand{\systemname}{MIAT}
\title{\LARGE \bf
MIAT: Maneuver-Intention-Aware Transformer for Spatio-Temporal Trajectory Prediction
}

\author{Chandra Raskoti$^{1}$, Iftekharul Islam$^{1}$, Xuan Wang$^{2}$, and Weizi Li$^{1}$ 
\thanks{$^{1}$Chandra Raskoti, Iftekharul Islam, and Weizi Li are with Min H. Kao Department of Electrical Engineering and Computer Science at University
of Tennessee, Knoxville, TN, USA
{\tt\small craskoti@vols.utk.edu, mislam73@vols.utk.edu, weizili@utk.edu}}%
\thanks{$^{2}$Xuan Wang is with Department of Electrical and Computer Engineering at George Mason University, Fairfax, VA, USA {\tt\small xwang64@gmu.edu}}%
}
\usepackage{graphicx}
\usepackage{amsmath}    
\usepackage{amssymb}
\usepackage{booktabs}
\usepackage{multirow}
\usepackage{csquotes}
\usepackage{adjustbox} 
\usepackage{subcaption} 

\usepackage[dvipsnames]{xcolor}

\usepackage{url}

\usepackage{algorithm}
\usepackage{algpseudocode}

\begin{document}

\maketitle
\thispagestyle{empty}
\pagestyle{empty}

\begin{abstract}

Accurate vehicle trajectory prediction is critical for safe and efficient autonomous driving, especially in mixed traffic environments when both human-driven and autonomous vehicles co-exist. However, uncertainties introduced by inherent driving behaviors---such as acceleration, deceleration, and left and right maneuvers---pose significant challenges for reliable trajectory prediction. We introduce a Maneuver-Intention-Aware Transformer (MIAT) architecture, which integrates a maneuver intention awareness control mechanism with spatiotemporal interaction modeling to enhance long-horizon trajectory predictions. We systematically investigate the impact of varying awareness of maneuver intention on both short- and long-horizon trajectory predictions. Evaluated on the real-world NGSIM dataset and benchmarked against various transformer- and LSTM-based methods, our approach achieves an improvement of up to 4.7\% in short-horizon predictions and a 1.6\% in long-horizon predictions compared to other intention-aware benchmark methods. Moreover, by leveraging intention awareness control mechanism, MIAT realizes an 11.1\% performance boost in long-horizon predictions, with a modest drop in short-horizon performance. The source code and datasets are available at https://github.com/cpraskoti/MIAT.

\end{abstract}

\section{Introduction}



\begin{figure}[t!]
    \centering
    \includegraphics[width=1.0\columnwidth]{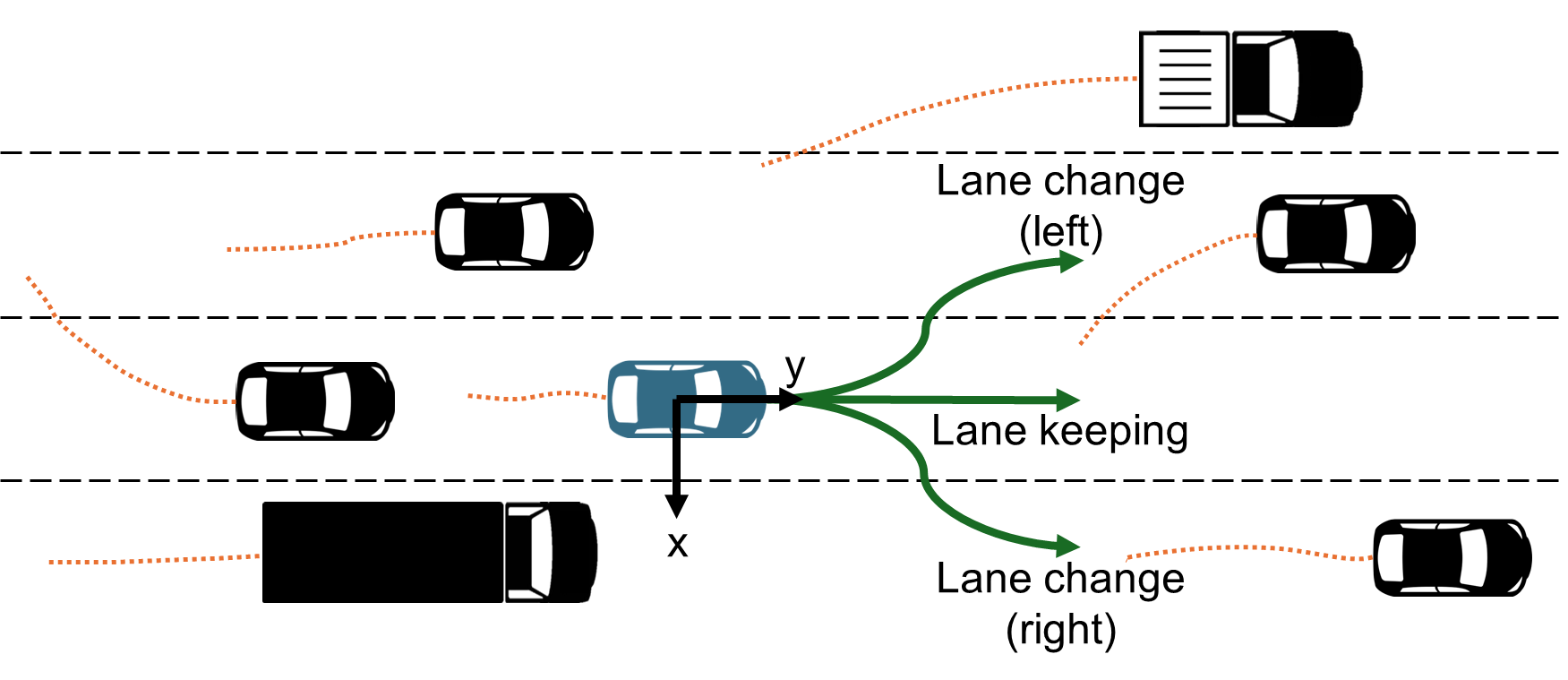} 
    \caption{\small{Illustration of trajectory prediction challenge. The blue vehicle (center) has multiple plausible future trajectories depending on different maneuver intentions: lane keeping (middle), changing to the left lane (top), or changing to the right lane (bottom). Neighboring vehicles (black) and their historical trajectories (orange dotted lines) influence the blue vehicle's decision-making process. 
    }}
    \label{fig:traj_prediction} 
    \vspace{-1em}
\end{figure}

The increasing integration of autonomous vehicles into everyday traffic is creating complex, mixed environments where human-driven and automated vehicles coexist~\cite{Chen2020AFI, wang2024learning, islam2024heterogeneous, poudel2024endurl}. The ability to accurately predict the future trajectories of surrounding vehicles is a fundamental requirement for safe and efficient autonomous navigation~\cite{houenou2013vehicle}. The core challenge lies in modeling the inherent uncertainties of human driving, which is characterized by a wide range of behaviors and intricate interactions among multiple agents. As illustrated in Fig.~\ref{fig:traj_prediction}, a single vehicle may have several plausible future paths corresponding to different high-level maneuvers, such as maintaining its lane or initiating a lane change. An effective prediction model must therefore account for this multi-modal nature of driving behavior.

Early deep-learning-based approaches for trajectory prediction often combine Long Short-Term Memory (LSTM) to capture the temporal evolution of vehicle states and social pooling or convolutional layers to model spatial interactions among neighboring agents~\cite{Alahi_2016_CVPR, deo2018convolutionalsocialpoolingvehicle, 9349962}. Despite their effectiveness in modeling local interactions, these methods are fundamentally limited by their recurrent backbone. The sequential processing nature of LSTMs makes it difficult to capture long-range dependencies efficiently, leading to error accumulation in long-horizon forecasts~\cite{xin2018intention}. Over time, Transformer~\cite{vaswani2023attentionneed} based architectures prove to be a compelling alternative to these prior models, using self-attention mechanisms to model complex spatio-temporal interactions across all agents and time steps simultaneously~\cite{Gao_2020_CVPR, Yuan_2021_ICCV}. While these models have significantly improved prediction accuracy, they typically learn high-level driving intentions implicitly. This lack of explicit maneuver reasoning can limit their performance and interpretability, especially in scenarios requiring consistent long-term forecasts where a driver's underlying intention is the primary factor guiding their path.

To address this limitation, other researchers have focused on explicitly incorporating maneuver awareness into prediction models via a two-stage pipeline: first classifying a driver's discrete maneuver such as lane keeping, left-right lane change and then generating a trajectory conditioned on that classification~\cite{chen2022intention, gao2023dual}. Other methods predict a sparse set of target goals corresponding to different maneuvers and generate trajectories toward them~\cite{tnt-pmlr-v155-zhao21b}. While effective, these methods often treat maneuver classification and trajectory generation as decoupled tasks. This separation can be suboptimal, as the balance between optimizing for short-term positional accuracy and long-term maneuver consistency is typically fixed and cannot be dynamically adjusted to prioritize long-horizon performance.

We introduce the Maneuver-Intention-Aware Transformer (\systemname), a unified architecture that bridges this gap by tightly coupling maneuver recognition and trajectory regression within a single end-to-end framework. Rather than employing a decoupled, two-stage process, \systemname integrates maneuver awareness directly into the learning objective of a Transformer-based model. Our central innovation is a tunable intention awareness control mechanism, which allows to explicitly control the trade-off between minimizing trajectory prediction error and correctly classifying the underlying maneuver. By adjusting the weight of the maneuver classification loss, our model can be configured to prioritize long-term strategic consistency, a key factor for reliable long-horizon forecasting.
Our contributions are as follows.

\begin{itemize} 


    \item We propose \systemname, a unified Transformer-based architecture that integrates a novel maneuver intention awareness and control mechanism. Our design jointly optimizes for trajectory accuracy and maneuver classification within a single framework to improve long-horizon prediction.


    \item Our tunable loss control mechanism explicitly controls the model's focus on long-term maneuver consistency versus short-term positional accuracy, which yields significant improvements in long-horizon performance (5-second horizon), achieving an 11.1\% improvement. 
    
    \item Our extensive experiments on the real-world NGSIM dataset demonstrate that \systemname~outperforms several benchmarks by up to 4.7\% for short-horizon and 1.6\% for long-horizon prediction.
\end{itemize}



\section{Related Work}
\label{section:related}
\subsection{Spatio-Temporal Models}
The prediction of vehicle trajectories requires capturing both the temporal evolution of a vehicle's state and its complex interactions with surrounding agents. Early studies often rely on classical probabilistic and physics-kinematic principles. Physics-based models leverage acceleration, yaw rate, and road friction to predict short-horizon trajectories but ignore driver intent and interactions with neighboring vehicles~\cite{rajamani2006vehicle}. To address this challenge, maneuver-based models integrate driver actions, such as lane-changing under kinematic constraints. For instance, maneuver recognition is combined with motion models~\cite{houenou2013vehicle}, while Bayesian frameworks are used to assess prediction reliability~\cite{schreier2014bayesian}. 
However, physics-based models are fundamentally limited by their reliance on rigid physical rules, which usually fail to capture the complex and unpredictable nature of driver behavior, thus restricting their accuracy to short-term predictions~\cite{huang2022survey}.

Early adoption of deep learning to trajectory prediction, particularly through the combination of LSTMs and Convolutional Networks, enabled the explicit modeling of complex spatio-temporal interactions. The seminal work on Social-LSTM~\cite{Alahi_2016_CVPR} introduced a ``social pooling'' layer to aggregate hidden states from neighboring agents, allowing the model to learn socially-aware behaviors. This concept was advanced by models such as CS-LSTM~\cite{deo2018convolutionalsocialpoolingvehicle}, which employs a convolutional grid to capture spatial proximity more effectively, and STA-LSTM~\cite{9349962}, which uses spatio-temporal attention to weigh the importance of different neighbors over time.

To address the inherent multi-modality of future trajectories, generative models are introduced. Social-GAN~\cite{gupta2018socialgansociallyacceptable} leverages a Generative Adversarial Network (GAN) to produce a diverse set of realistic and socially plausible paths, while others such as DESIRE~\cite{Lee_2017_CVPR} uses Variational Autoencoders (VAEs) to model a probabilistic distribution over multiple futures.

Despite their success, RNN-based architectures struggle to model complex, long-range spatio-temporal interactions due to their sequential nature. This development leads to the adoption of Transformer architectures~\cite{vaswani2023attentionneed}, whose self-attention mechanisms can capture global correlations across time and agents simultaneously. Seminal work including VectorNet~\cite{Gao_2020_CVPR} and AgentFormer~\cite{Yuan_2021_ICCV} demonstrate the power of attention-based models. VectorNet encodes the scene as a graph of vectors, using self-attention to learn rich context-aware representations, while AgentFormer models inter-agent dependencies to generate consistent joint predictions. 
While these models effectively capture complex interactions, they differ from MIAT by learning high-level maneuvers implicitly. This lack of an explicit mechanism for maneuver awareness and control constrains their performance in long-horizon prediction.


\subsection{Maneuver Aware Models}
Building on the strong spatio-temporal modeling capabilities of modern architectures, a distinct line of research focuses on explicitly incorporating high-level driving maneuvers to improve prediction accuracy and interpretability. This is achieved through two main strategies: goal-oriented prediction and maneuver-conditioned generation.

Goal-oriented approaches frame trajectory prediction as a two-step process: first predicting a set of plausible future goals or waypoints, and then generating a full trajectory conditioned on a selected goal. MultiPath~\cite{chai2019multipath} anchors this idea by predicting both a probability distribution over a fixed set of trajectory anchors to model intent and a continuous refinement for each anchor to model control, yielding an efficient, multi-modal forecast. Recent methods such as TNT~\cite{tnt-pmlr-v155-zhao21b} predict a sparse set of target waypoints representing distinct maneuver outcomes and then generate trajectories toward them. Similarly, LaneGCN~\cite{liang2020learninglanegraphrepresentations} leverages HD map information to propose multiple plausible paths along the lane graph, with each path corresponding to a specific maneuver.


Another dominant strategy involves explicitly classifying the driver's maneuver and using this classification to guide a trajectory generator. This two-stage pipeline first predicts a distribution over discrete maneuvers such as lane keeping, left-right lane change and conditions an LSTM or Transformer-based decoder on this latent variable. While these methods provide explicit reasoning, they often rely on architectures where the classification and regression tasks are decoupled, which can be suboptimal~\cite{chen2022intention, gao2023dual}.

In contrast to these approaches, \systemname~introduces a novel, tightly coupled framework. Instead of treating maneuver prediction as a separate classification task or a target selection problem, we integrate maneuver awareness directly into the learning objective of a unified Transformer architecture. Our key contribution is a tunable loss weighting mechanism that allows the model to balance the importance of short-term positional accuracy against long-term maneuver recognition. This design provides a distinct advantage in explicitly improving long-horizon predictions, which is a direct benefit of emphasizing the maneuver awareness component during training. While existing Transformer-based studies have established the architecture's general effectiveness, \systemname~is the first to systematically leverage a tunable, maneuver-aware objective within a Transformer coupled framework for trajectory prediction.

\section{Methodology}
\label{section:methodology}



\begin{figure}[t!]
    \centering
    \includegraphics[width=0.48\textwidth]{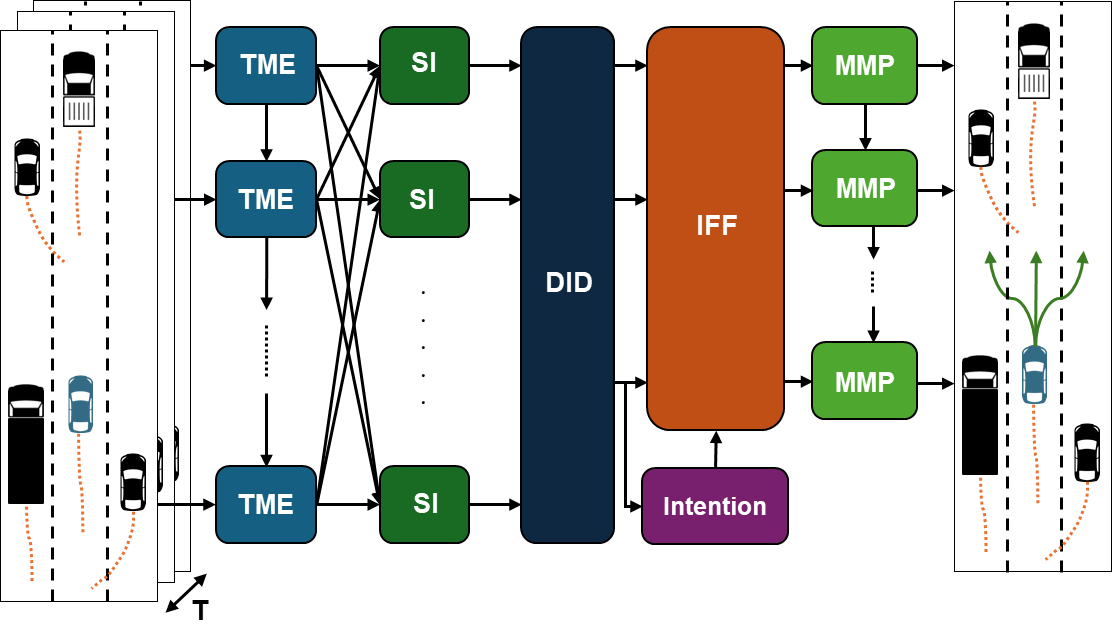}
    \caption{\small{\systemname's architecture. \systemname~processes multi-vehicle historical trajectories (left) through parallel Transformer Motion Encoders (TME) and Social Interaction (SI) module for each timestamp. The Dynamic Interaction Dependency (DID) module captures how inter-vehicle relationships evolve. Combined with driving intention predictions, the Intention-Specific Feature Fusion (IFF) module selectively weighs temporal features for each possible maneuver. Finally, multiple Multi-Modal Prediction (MMP) module generate diverse trajectory predictions (right) corresponding to different possible driving intentions.}}
    \label{fig:model_architecture}
\vspace{-1em}
\end{figure}

\subsection{Problem Formulation}
The trajectory prediction problem is formulated as the task of forecasting future positions of a target vehicle based on its historical motion and interactions with surrounding vehicles. $X_t = \{x^0_t, x^1_t, \ldots, x^N_t\}$ represent the states of \(N+1\) vehicles at timestamp \(t\), where each \(x^0_t\) and  \(x^i_t\)$\small{(i\geqslant1)}$ includes features of the ego and neighboring vehicle such as position, velocity, and acceleration, respectively. The goal is to predict the future trajectory $Y = \{y^0_{T+1}, \ldots, y^0_{T+F}\}$
where \(y^0_{T+f}\) denotes the coordinates of the target vehicle at a future time step. Our model learns the probability distribution \(P(Y|X)\) over future trajectories by explicitly incorporating maneuver awareness into the prediction process. 

\subsection{Model Architecture}

\systemname~extends the STDAN architecture~\cite{chen2022intention} to more effectively capture long-range dependencies. As depicted in Fig.\ref{fig:model_architecture}, the \systemname~architecture is composed of several key components, each designed to process distinct aspects of spatio-temporal dynamics and driving maneuvers.

\subsubsection{Transformer Motion Encoder}
The Transformer Motion Encoder (TME) is responsible for capturing temporal dynamics in historical trajectory data of both ego and neighboring vehicles. It first projects the raw trajectory features into a latent space where temporal dependencies are more effectively captured. We use a Transformer-based encoding mechanism that allows for parallel processing and better modeling of long-range complex dependencies. The module learns an embedding of each vehicle’s state at each timestamp, capturing both immediate and long-term motions. 
For the target vehicle, let \(\mathbf{x}_t \in \mathbb{R}^{F}\) denote the raw features at time \(t\). An intermediate representation is computed as $
e_t = \sigma\left(\mathbf{x}_t\,W_e\right)$, where \(\sigma(\cdot)\) is a nonlinear activation (e.g., LeakyReLU); this is then projected to the Transformer latent dimension $z_t = e_t\,W_p$, 
where $W_p$ is projection embedding parameters and the sequence \(\{z_1, \dots, z_T\}\) is processed by a Transformer encoder:
\[
H^0 = \text{TransformerEncoder}\bigl(\{z_t\}_{t=1}^T\bigr).
\]
A similar encoding is applied to all neighboring vehicles’ trajectories, whose outputs are aligned into a spatial grid for subsequent interaction modeling.

\subsubsection{Spatial Interaction}
To account for the influence of surrounding vehicles, the Social Interaction module (SI) aggregates features from the neighbors. For a given time \(t\), let \(h_t^0 \in \mathbb{R}^{d}\) be the encoded feature of the target vehicle and let \(S_t \in \mathbb{R}^{G \times d}\) represent the grid of neighboring features. Simplified key, query, and value are computed as 
\[
q_t = h_t^0\,W_q, \quad K_t = S_t\,W_k, \quad V_t = S_t\,W_v.
\]
The attention weights are then calculated as
\[
\alpha_t = \text{softmax}\!\left(\frac{q_t\,K_t^\top}{\sqrt{d}}\right).
\]
The aggregated social feature is $h_t^{\text{social}} = \alpha_t\,V_t.$
A Gated Linear Unit (GLU) is applied to \(h_t^{\text{social}}\) and combined with the original target feature via a residual connection:
\[
\tilde{h}_t = \text{LayerNorm}\Bigl(h_t^0 + \text{GLU}\bigl(h_t^{\text{social}}\bigr)\Bigr).
\]

\subsubsection{Dynamic Interaction Dependency}
The SI module calculates social interaction feature for each timestamp independently, without accounting for temporal correlations among social representations. However, the social dependency at different timestamps could be temporally
correlated. Thus, the Dynamic Interaction Dependency module (DID) is used to capture the relationship between the social interaction representations across different timestamps using multi-head self-attention mechanism: 
\begin{align*}
Q &= \tilde{H}\,W_{q_t}, \quad K = \tilde{H}\,W_{k_t}, \quad V = \tilde{H}\,W_{v_t},
\end{align*}
where \(\tilde{H}\) stacks \(\tilde{h}_t\) over time. The temporal attention is computed as
\[
\beta = \text{softmax}\!\left(\frac{Q\,K^\top}{\sqrt{d}}\right).
\]
The temporally aggregated feature is $H^{\text{temp}} = \beta\,V.$ 
A subsequent GLU and residual connection yield the final spatio-temporal encoding:
\[
\tilde{H} = \text{LayerNorm}\Bigl(\tilde{H} + \text{GLU}\bigl(H^{\text{temp}}\bigr)\Bigr).
\]
Both TME and DID modules capture temporal dependencies.
The distinction is that ME extracts temporal relations from raw trajectory data, while DID captures the temporal correlations within social interaction. 

\subsubsection{Intention-Specific Feature Fusion and Prediction}
Inherent driving characteristics such as acceleration, deceleration, and left and right maneuvers contribute to the uncertainty of vehicle trajectories. 
We classify these maneuvers into six types: lateral--lane keeping (LK), lane change left (CLL), lane change right (CLR), and longitudinal--acceleration (ACC), deceleration (DEC), constant speed (CON). 
The Intention-Specific Feature Fusion module (IFF) addresses the uncertainty due to these maneuvers by incorporating an intention recognition component that estimates the probability of various driving
maneuvers. A maneuver state vector is computed as $r = \sigma\left(\tilde{h}_T\,W_r\right).$ 
This state vector is split into lateral and longitudinal components:
\begin{align*}
h_{\text{la}} &= r\,W_{\text{la}}, \quad P(\text{la}) = \text{softmax}(h_{\text{la}}), \\
h_{\text{lo}} &= r\,W_{\text{lo}}, \quad P(\text{lo}) = \text{softmax}(h_{\text{lo}}).
\end{align*}
These probabilities condition a learned mapping that fuses the spatio-temporal encoding:
\[
d = f\bigl(\tilde{H},\,P(\text{la}),\,P(\text{lo})\bigr),
\]
where \(f(\cdot)\) denotes the mapping operation implemented via soft attention over the encoded features. The fused feature \(d\) is then input to a Transformer-encoder layer and hidden state is computed for each timestep \(t'\) (with \(t' = T+1, \dots, T+F\)).
\[
h_{t'} = \text{TransformerDecoder}\bigl(d\bigr),
\]
The Transformer layer captures the combined socio-temporal context and maneuver awareness.
Lastly, the hidden state $h_{t'}$ is fed to MLP that maps it to the parameters of a bivariate Gaussian distribution:
\[
\theta_{t'} = \{\mu_{t',x}, \mu_{t',y}, \sigma_{t',x}, \sigma_{t',y}\} = \text{MLP}(h_{t'}), 
\]
where \(\mu_{t',x}\) and \(\mu_{t',y}\) are the predicted means for the coordinates, \(\sigma_{t',x}\) and \(\sigma_{t',y}\) are the standard deviations. This final step converts the encoded features into a probabilistic prediction of future vehicle positions.

\subsection{Loss Function}
The overall training objective combines the trajectory prediction loss and the maneuver classification loss:
\[
L = L_{\text{traj}} + \lambda \cdot L_{\text{maneuver}},
\]
where \(L_{\text{traj}}\) is initially mean square loss and in later epochs negative log-likelihood of the ground truth trajectory under the predicted Gaussian distribution is used. \(L_{\text{maneuver}}\) is the cross-entropy loss between the predicted and true maneuver labels. 
To enhance optimization, a two-stage training strategy is employed where, initially, the model minimizes a simpler Mean Squared Error (MSE) loss for a few epochs, establishing a solid baseline through smooth gradients. After this warm-up phase, the more expressive Negative Log Likelihood (NLL) loss is used, enabling the model to capture the probabilistic uncertainty in trajectory predictions.

\subsection{Benchmarks} 

We evaluate the performance of our proposed method against three distinct classes of baseline methods: attention-based Transformer architectures, models built on recurrent networks for spatio-temporal modeling,  and other specialized methods.

\subsubsection{Attention-based Spatio-Temporal Architecture}

STDAN~\cite{chen2022intention} employs a Transformer-based architecture to model sociotemporal dependencies in vehicle interactions.  
SIT~\cite{ijgi11020079SIT} utilizes multi-head self-attention to capture both spatial and temporal dependencies.  
Vanilla Transformer (Vanilla TF) is a Transformer encoder-decoder baseline without maneuver awareness, used for ablation studies.  

\subsubsection{Spatio-Temporal Recurrent Architecture}

DLM~\cite{dlm} integrates an occupancy and risk map into an LSTM-based encoder-decoder for interaction-aware predictions.  
STA-LSTM~\cite{9349962} applies spatial-temporal attention to selectively weight vehicle interactions over time.  
NLS-LSTM~\cite{10.1109/IVS.2019.8813829} merges local and non-local operations to model inter-vehicle dependencies.  
CS-LSTM~\cite{deo2018convolutionalsocialpoolingvehicle} combines convolutional pooling with LSTMs to model spatial interactions.  
S-LSTM~\cite{Alahi_2016_CVPR} encodes trajectories using LSTMs with a social pooling mechanism.  

\subsubsection{Hybrid Approaches}

PiP~\cite{PIP} conditions trajectory forecasts on candidate motion plans, linking prediction with planning.  
DSCAN~\cite{ijgi10050336} integrates attention for dynamic vehicle prioritization and static context modeling.  
S-GAN~\cite{gupta2018socialgansociallyacceptable} introduces adversarial learning to generate socially plausible multi-modal trajectories.

\section{Experiment}
\label{section:experimenatalsetup}

\subsection{Dataset}
\begin{figure}[t!]
\centering
\includegraphics[width=0.48\textwidth]{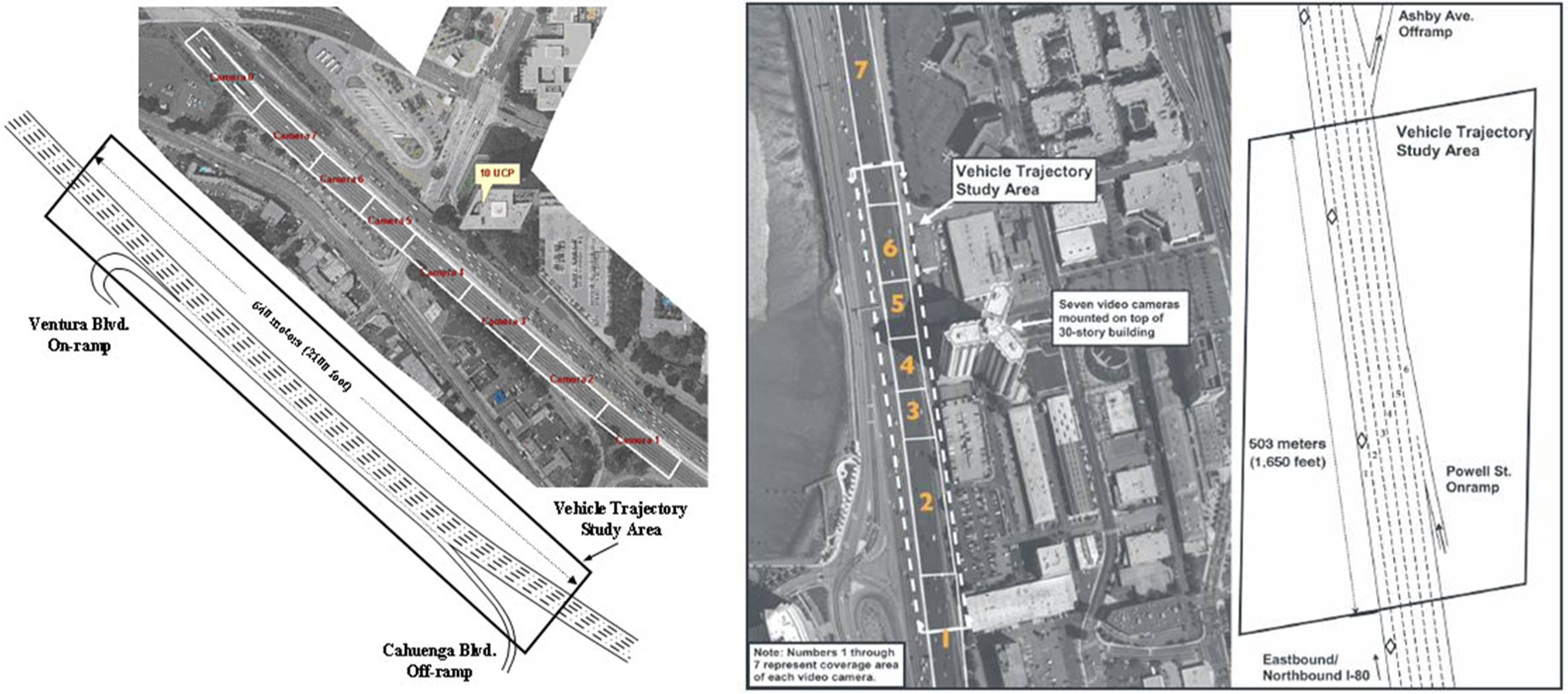}
\caption{\small{Aerial Overview of the NGSIM study area US 101 (left) and I‑80 (right) in relation to the building from which the digital video cameras are mounted and the coverage area for each of the eight cameras~\cite{NGSIM2016}.
}}
\label{fig:NGSIM_study_area}
\vspace{-1.5em}
\end{figure}

\systemname~is evaluated using the NGSIM dataset (Fig.\ref{fig:NGSIM_study_area}), which contains vehicle trajectory data collected from the US-101 and I-80 freeways~\cite{NGSIM2016}. The dataset provides detailed vehicle trajectories including positions, velocities, and lane information.
Preprocessing consists of several steps. First, lane IDs are standardized with a maximum value capped at six to ensure consistency across the dataset. Maneuver detection is performed using 40-frame windows (approximately 4 seconds) to capture complete vehicle movements. Subsequently, edge cases where vehicles lack sufficient trajectory history are removed to ensure reliable model training. Lastly, the data is split into training (70\%), validation (10\%), and test (20\%) sets while maintaining unique vehicle IDs across the splits. The final dataset contains 5,922,867 training, 859,769 validation, and 1,505,756 test entries.




\subsection{Training Setup}
We implement our model using the Transformer architecture with specific configurations for trajectory prediction. Our model utilizes a single encoder layer with eight attention heads, totaling 729,979 parameters, and processes inputs on a $3 \times 13$ spatial grid. For training, we employ the Adam optimizer with a learning rate of $1 \times 10^{-4}$. All experiments are conducted on a system with Intel i9-14900K CPU and Nvidia GTX 4090 GPU. These parameters are chosen to ensure computational efficiency while maintaining a fair comparison with baseline methods.

\subsection{Evaluation}
We evaluate the trajectory forecasting performance using Root Mean Square Error (RMSE):
\begin{equation*}
RMSE = \sqrt{\frac{1}{N} \sum_{i=1}^{N} \sum_{k=1}^{K} \left[ (\hat{x}_{i,k} - x_{i,k})^2 + (\hat{y}_{i,k} - y_{i,k})^2 \right]},
\end{equation*}
where \(N\) is the number of samples, \(K\) is the prediction horizon, and \((\hat{x}_{i,k}, \hat{y}_{i,k})\) and \((x_{i,k}, y_{i,k})\) represent predicted and actual coordinates, respectively.

\section{Results}
\label{section:result}



We evaluate \systemname~against models different spatio temporal and maneuver aware models closely aligned with our approach for trajectory prediction and analyze the impact of our design choices in longer forecast horizon through an ablation study.

\subsection{Comparative Analysis}

Table~\ref{tab:benchmark_table} presents the performance comparison between our method with no maneuver loss scaling (\systemname-NoScale), 200x scaling (\systemname-200x), and several baselines methods across prediction horizons ranging from 1 to 5 seconds. For comparative analysis, we include Transformer-based approaches (STDAN~\cite{chen2022intention}, SIT~\cite{ijgi11020079SIT}), LSTM-based methods (DLM~\cite{dlm}, STA-LSTM~\cite{9349962}, S-GAN~\cite{gupta2018socialgansociallyacceptable}, NLS-LSTM~\cite{10.1109/IVS.2019.8813829}, CS-LSTM~\cite{deo2018convolutionalsocialpoolingvehicle}, S-LSTM~\cite{Alahi_2016_CVPR}), and Hybrid models (PiP~\cite{PIP}, DSCAN~\cite{ijgi10050336}).  
Additionally, we implement a vanilla Transformer encoder-decoder model without our maneuver-aware component for the ablation study.

\begin{table}[h]

\centering
\setlength{\tabcolsep}{3.8pt} 

\begin{adjustbox}{max width=\columnwidth}
\begin{tabular}{lccccc} 

\toprule
\multicolumn{1}{c}{\multirow{2}{*}{}} & \multicolumn{5}{c}{Prediction Horizon} \\
\cmidrule(lr){2-6}
                                                  Model & 1~s & 2~s & 3~s & 4~s & 5~s \\
\midrule
$\shortstack{\text{\systemname-NoScale}}$                        & $\textbf{0.40}_\textbf{{\text{\tiny (+4.7\%)}}}$ & $0.98_{\text{\tiny (+2.9\%)}}$ & $\textbf{1.65}_{\text{\tiny \textbf{(+2.3\%)}}}$ & $\textbf{2.52}_{\text{\tiny \textbf{(+1.5\%)}}}$ & $\textbf{3.61}_{\text{\tiny \textbf{(+1.6\%)}}}$ \\
$\shortstack{\text{\systemname-200x}}$      & $0.44_{\text{\tiny (-4.7\%)}}$ & $0.98_{\text{\tiny (+2.9\%)}}$  & $\textbf{1.58}_{\text{\tiny \textbf{(+6.5\%)}}}$  & $\textbf{2.31}_{\text{\tiny \textbf{(+9.7\%)}}}$  & $\textbf{3.26}_{\text{\tiny \textbf{(+11.1\%)}}}$  \\
Vanilla TF                                         & 0.61 & 1.31 & 2.17 & 3.23 & 4.57 \\
\midrule
STDAN~\cite{chen2022intention}                                              & 0.42 & 1.01 & 1.69 & 2.56 & 3.67 \\
DLM~\cite{dlm}                                                & 0.41 & \textbf{0.95} & 1.72 & 2.64 & 3.87 \\
PiP~\cite{PIP}                                                & 0.55 & 1.18 & 1.94 & 2.88 & 4.04 \\
SIT~\cite{ijgi11020079SIT}                                         & 0.58 & 1.23 & 1.99 & 2.96 & 4.05 \\
DSCAN~\cite{ijgi10050336}                                         & 0.57 & 1.25 & 2.03 & 2.98 & 4.13 \\
STA-LSTM~\cite{9349962}                                           & 0.59 & 1.25 & 2.03 & 3.03 & 4.28 \\
S-GAN~\cite{gupta2018socialgansociallyacceptable}                                              & 0.57 & 1.32 & 2.22 & 3.26 & 4.40 \\
NLS-LSTM~\cite{10.1109/IVS.2019.8813829}                                           & 0.56 & 1.22 & 2.02 & 3.03 & 4.30 \\
CS-LSTM~\cite{deo2018convolutionalsocialpoolingvehicle}                                            & 0.61 & 1.27 & 2.09 & 3.10 & 4.37 \\
S-LSTM~\cite{Alahi_2016_CVPR}                                             & 0.65 & 1.31 & 2.16 & 3.25 & 4.55 \\
\bottomrule
\end{tabular}
\end{adjustbox}
\caption{\small{RMSE (m) comparison for trajectory prediction models. Our methods \systemname-NoScale and \systemname-200x are evaluated against other benchmarks. Subscripts are the \% RMSE improvements in comparison to the best performing benchmark model STDAN. \systemname-NoScale achieves the lowest RMSE at 1, 3, 4, and 5~s compared to STDAN. \systemname-200x outperforms all baselines over longer horizons (3--5~s), achieving up to an 11.1\% RMSE improvement over STDAN at 5~s. Although DLM performs the best at 2~s, our models still have a 2.9\% improvement over STDAN.}}
\label{tab:benchmark_table}
\end{table}

\systemname-NoScale achieves the lowest prediction error at the 1-second horizon (RMSE = 0.40), outperforming all baselines including STDAN (0.42) and DLM (0.41). For long-horizon predictions (3--5 seconds),  \systemname-200x demonstrates consistent superiority, achieving RMSE improvements by 6.5\%, 9.7\%, and 11.1\% for the 3, 4, and 5-second horizons, respectively, compared to STDAN. The improvements increases progressively as the prediction horizon increases.

Several findings emerge from these results. First, all models exhibit an increase in error with longer prediction horizons, which aligns with the inherent difficulty of forecasting vehicle trajectories as the temporal distance increases. However, error accumulation rate varies among models. \systemname-NoScale and \systemname-200x have reduced RMSE accumulation rate by approximately 1.2\% and 1.3\% per second (compared to STDAN), respectively. 
There exists a trade-off in performance between short and long horizon prediction without scaling vs. with scaling the maneuver loss. Scaling maneuver loss yields up to 11.1\% improvements compared to just 1.6\% in without scaling at long horizon with the sacrifice of 4.7\% in short horizon. This shows the potential importance of maneuver awareness in long horizon predictions.

Second, the substantial performance gap between our \systemname~variants and the vanilla Transformer underscores the critical importance of the maneuver-aware component. Without explicit modeling of driving intentions and maneuver-specific feature integrations, self-attention mechanisms of Transformers alone fail to capture the nuanced patterns of vehicle movements. 


Third, the consistent underperformance of LSTM-based methods relative to attention-based approaches, especially at longer horizons, highlights the limitations of sequential processing in capturing complex, long-range spatio-temporal dependencies. 

\begin{figure*}[t!]
\centering
\includegraphics[width=\textwidth]{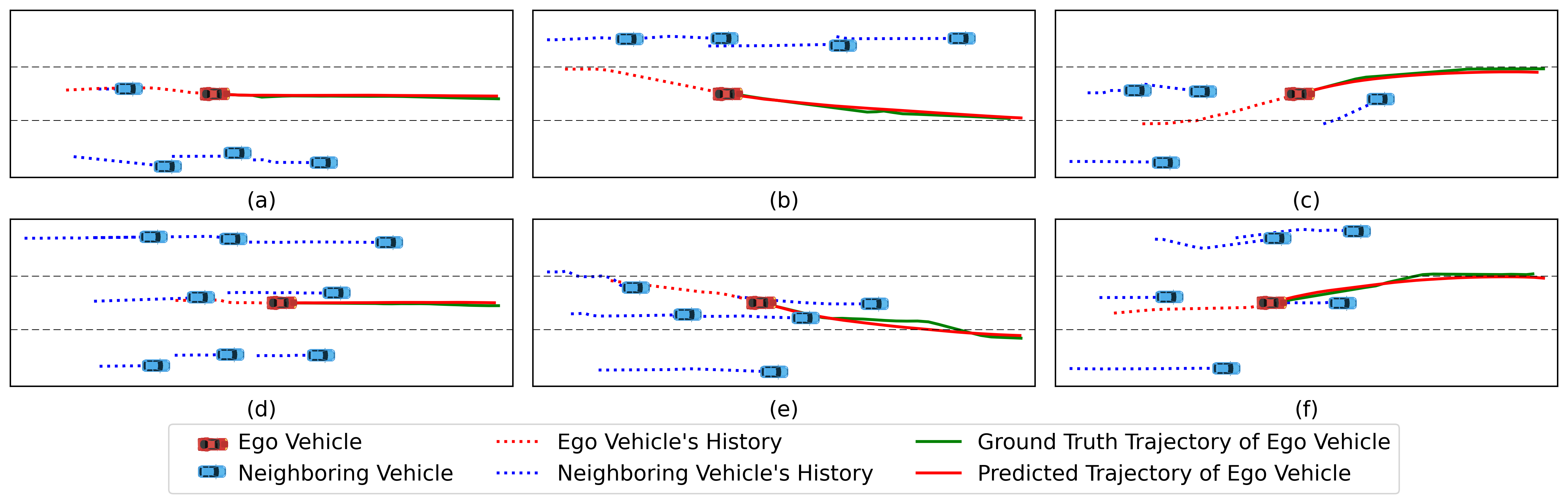}
\caption{\small{Historical, predicted, and ground truth trajectories for the ego and neighboring vehicles under different lane maneuvers. The top row (panels a, b, c) is for light traffic density, while the bottom row (panels d, e, f) is for heavy traffic density. The left column (a, d) shows lane keeping, the middle column (b, e) shows right maneuver, and the right column (c, f) shows left maneuver.}
}
\label{fig:trajectory_illustrations}

\end{figure*}

\subsection{Qualitative Trajectory Analysis}
Trajectory predictions across three common driving maneuvers: lane keeping, left  maneuver, and right maneuver in both light and dense traffic are shown in Fig.~\ref{fig:trajectory_illustrations}. The visualization includes historical trajectories (red dotted lines), neighboring vehicle paths (blue dotted lines), ground truth trajectories (green solid lines), and \systemname's predictions (red solid lines).

In lane-keeping scenarios (Fig.~\ref{fig:trajectory_illustrations} a and d), \systemname ~accurately maintains the vehicle’s lane trajectory both in high and low traffic density while predicting minor lateral adjustments necessary for proper positioning. As the prediction horizon increases, compounding errors lead to a growing discrepancy between the ground truth and the predicted trajectories, as evident in the trailing parts of the prediction plots. In lane change maneuvers (Fig.\ref{fig:trajectory_illustrations} middle and right columns), \systemname ~captures both the initiation and execution phases for lateral maneuvers. Notably, in complex scenarios with multiple neighboring vehicles, the \systemname ~effectively prioritizes those most relevant interactions based on their relative positions and motion patterns. This is evident in Fig.~\ref{fig:trajectory_illustrations} (f), where the model accurately predicts a delayed lane change execution due to vehicles in the target lane.

\subsection{Ablation Study}

\begin{figure}[t!]
\centering
\includegraphics[width=8cm]{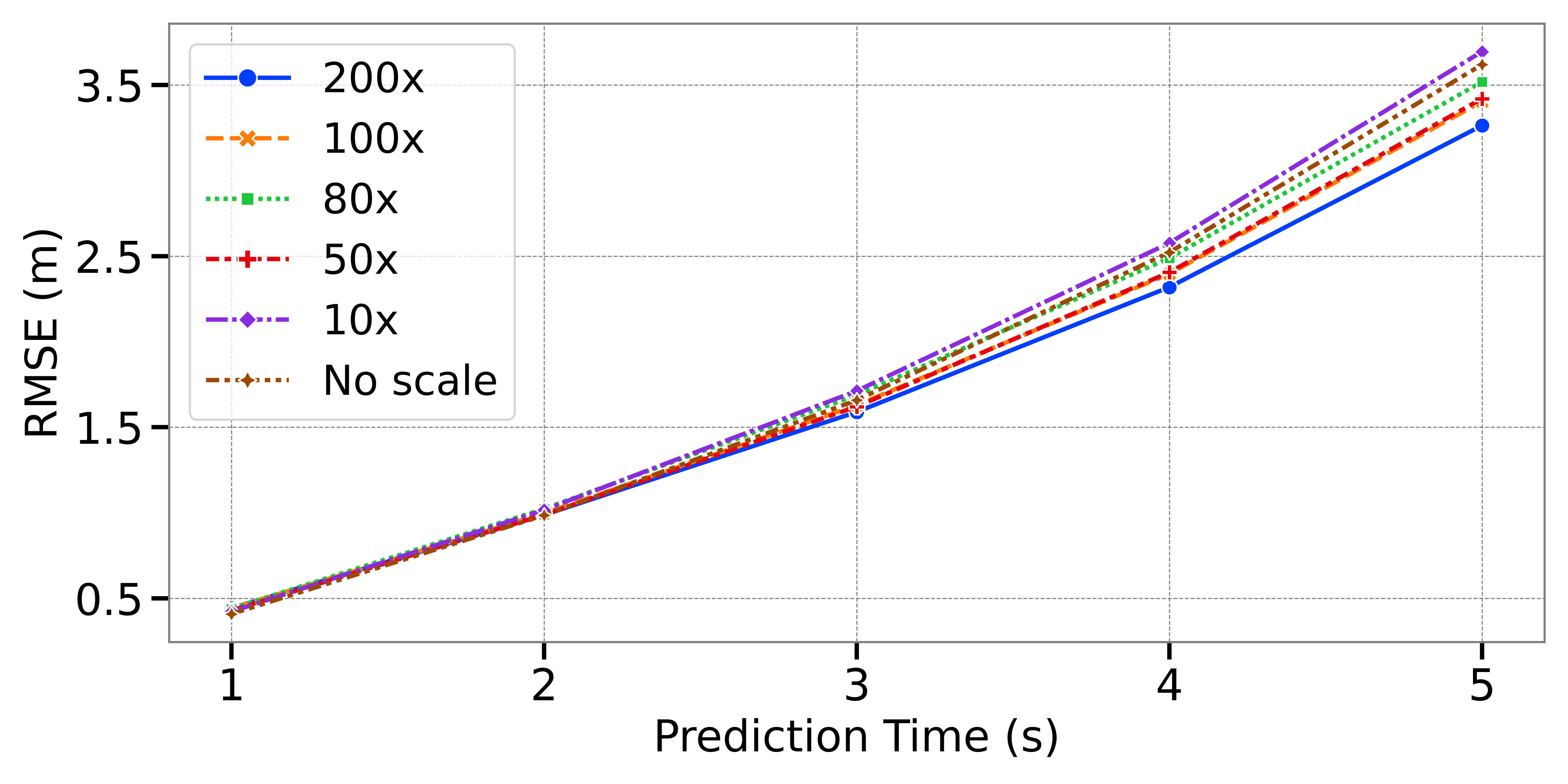}
\caption{\small{Ablation study comparing the effect of different maneuver loss weightings. The results show that a moderate scaling factor (10x) degrades performance compared to no scaling, while higher scaling factors (50x, 80x, 100x, and 200x) reduce RMSE for longer forecast horizons (3--5~s). The higher the scaling factor, the lower the long-horizon error-with the lowest error achieved at 200x suggesting that a stronger emphasis on maneuver loss is crucial for accurately capturing long-term trajectory changes.}}
\label{fig:RMSE Comparison: No scaling vs. Scaled Maneuver Loss Weighting}
\end{figure}

Unlike previous studies such as STDAN, we used a weighted maneuver loss to study the effect of varying levels of maneuver awareness on trajectory prediction. 
To understand the impact of different weights on maneuver awareness, we conducted ablation study focusing on the maneuver loss weighting parameter $\lambda$ in our unified loss function. Fig.~\ref{fig:RMSE Comparison: No scaling vs. Scaled Maneuver Loss Weighting} illustrates the effect of different scaling factors (10x, 50x, 80x, 100x, 200x) compared to no scaling ($\lambda = 1$) on prediction accuracy over all forecast horizons.

Using a 10x scaling factor leads to performance degradation in both short- and long-horizon predictions, likely due to insufficient maneuver information in the network. In contrast, as the scaling factor is incrementally increased from 50x to our highest tested value of 200x, long-horizon prediction performance improves progressively, despite a slight degradation in short-horizon predictions. This suggests that maneuver trajectory information is critical for modeling extended driving behaviors in line with real-world dynamics.

\begin{table}[tb]
\centering
\setlength{\tabcolsep}{4pt} 
\begin{tabular}{l c c c c c}
\toprule
\multicolumn{1}{c}{\multirow{2}{*}{\shortstack{Scaling\\ Factor}}} & \multicolumn{5}{c}{Prediction Horizon} \\
\cline{2-6}
 & 1~s & 2~s & 3~s & 4~s & 5~s \\
\midrule
No scale & \textbf{0.40} & 0.98 & 1.65 & 2.52 & 3.61 \\
\hline
10x      & $0.42_{\text{\tiny (-5.0\%)}}$  & $1.01_{\text{\tiny (-3.0\%)}}$  & $1.71_{\text{\tiny (-3.6\%)}}$  & $2.57_{\text{\tiny (-1.9\%)}}$  & $3.69_{\text{\tiny (-2.2\%)}}$  \\
50x      & $0.42_{\text{\tiny (-5.0\%)}}$  & $0.98_{\text{\tiny (0\%)}}$  & $1.62_{\text{\tiny (+1.8\%)}}$  & $2.40_{\text{\tiny (+4.7\%)}}$  & $3.42_{\text{\tiny (+5.2\%)}}$  \\
80x      & $0.43_{\text{\tiny (-7.5\%)}}$  & $1.02_{\text{\tiny (-4.0\%)}}$  & $1.68_{\text{\tiny (-1.8\%)}}$  & $2.48_{\text{\tiny (+1.5\%)}}$  & $3.51_{\text{\tiny (+2.7\%)}}$  \\
100x     & $0.44_{\text{\tiny (-10.0\%)}}$ & $0.99_{\text{\tiny (-1.0\%)}}$  & $1.62_{\text{\tiny (+1.8\%)}}$  & $2.39_{\text{\tiny (+5.1\%)}}$  & $3.40_{\text{\tiny (+5.8\%)}}$  \\
200x     & $0.44_{\text{\tiny (-10.0\%)}}$ & $\textbf{0.98}_{\text{\tiny \textbf{(0\%)}}}$  & $\textbf{1.58}_{\text{\tiny \textbf{(+4.2\%)}}}$  & $\textbf{2.31}_{\text{\tiny \textbf{(+8.3\%)}}}$  & $\textbf{3.26}_{\text{\tiny \textbf{(+9.6\%)}}}$  \\
\bottomrule
\end{tabular}
\caption{
\small{Ablation study results comparing the impact of various weighting factors on maneuver loss for both long‐ and short‐horizon predictions. Performance does not improve with small factors (10x). However, as the weighting factor increases, the RMSE~(m) for a long prediction horizon (3--5~s) improves by up to 9.6\% at 5~s for 200x scaling compared to unscaled predictions, while the RMSE for short‐horizon predictions (1~s) degrades by 10\%.}}
\label{tab:scaling_predictions}
\end{table}

\section{Conclusion and Future Work}
\label{section:conclusion}



This work introduces two key advancements over prior intention-aware models. First, a tunable intention awareness control mechanism is presented. Unlike approaches that use a fixed weight loss for trajectory and maneuver objectives, this mechanism allows for a dynamic balance between trajectory loss and maneuver loss, effectively balancing short-term precision and long-term consistency. Notably, a deliberate 200 times scale of the maneuver loss yielded an 11.1\% better performance in 5-second horizon compared to STDAN, demonstrating a promising method for enhancing long-horizon forecasting.

Second, a Transformer architecture was integrated, replacing conventional LSTM backbones. This architectural shift enables \systemname~to effectively model long-range spatio-temporal dependencies, providing a robust foundation for an intention-aware learning strategy. This approach is crucial for achieving reliable performance over extended horizons, directly addressing the inherent limitations of traditional sequential models. Ultimately, \systemname~offers a robust framework and competitive methodology for real-time trajectory prediction systems, particularly in applications where long-horizon reliability is paramount.

Future work involves enhancing \systemname~by incorporating graph-based reasoning to leverage road topology. The framework will also be extended to generate multi-modal trajectory distributions to capture a wider range of plausible outcomes, and its robustness will be validated on larger, more complex datasets.

\section*{Acknowledgements}
This research is supported by NSF IIS-2153426. The
authors thank NVIDIA and the Tickle College of Engineering
at University of Tennessee, Knoxville for their support.

\bibliographystyle{unsrt}
\bibliography{references}

\end{document}